\documentclass[manuscript,nonacm]{acmart} 

\usepackage{macros}
\usepackage[english]{babel}
\usepackage{amsthm}
\usepackage{graphicx, array, blindtext}
\usepackage{dblfloatfix}
\theoremstyle{definition}

\usepackage{subcaption}
\usepackage{newfloat}
\usepackage{adjustbox}
\usepackage{listings}
\usepackage{xcolor}
\definecolor{mydarkblue}{rgb}{0,0.08,0.45}
\usepackage{cleveref}
\usepackage{tikz}
\usepackage{mathtools} 
\usepackage{amsfonts} %
\usepackage{pgfplots}
\pgfplotsset{compat=1.18}
\usetikzlibrary[pgfplots.groupplots]
\usepackage{wrapfig}
\usepackage{booktabs}

\usepackage[symbol]{footmisc}

\usetikzlibrary{arrows, automata, positioning}
\tikzset{auto, >=stealth}
\tikzset{every edge/.append style={shorten >= 1pt}}
\tikzset{
    main node/.style={circle,draw,minimum size=1cm,inner sep=0pt},
}

\DeclareCaptionStyle{ruled}{labelfont=normalfont,labelsep=colon,strut=off} 
\floatstyle{ruled}
\newfloat{listing}{tb}{lst}{}
\floatname{listing}{Listing}

\usepackage[font=small,skip=2pt]{caption}

\AtBeginDocument{%
  }

\setcopyright{none}
\acmConference[Anonymous Submission]{}{Year}{Location}
\acmBooktitle{}
\acmPrice{}
\acmDOI{}
\acmISBN{}
\renewcommand\footnotetextcopyrightpermission[1]{} 
\begin{document}

\title{Evaluating Human Trust in LLM-Based Planners: A Preliminary Study}

\author{Shenghui Chen}
\affiliation{%
  \institution{University of Texas at Austin}
  \city{Austin}
  \state{Texas}
  \country{USA}
}
\email{shenghui.chen@utexas.edu}

\author{Yunhao Yang}
\affiliation{%
  \institution{University of Texas at Austin}
  \city{Austin}
  \state{Texas}
  \country{USA}
}
\email{yunhaoyang234@utexas.edu}

\author{Kayla Boggess}
\affiliation{
  \institution{University of Virginia}
  \city{Charlottesville}
  \state{Virginia}
  \country{USA}
}
\email{kjb5we@virginia.edu}

\author{Seongkook Heo}
\affiliation{
  \institution{University of Virginia}
  \city{Charlottesville}
  \state{Virginia}
  \country{USA}
}
\email{seongkook@virginia.edu}

\author{Lu Feng}
\affiliation{
  \institution{University of Virginia}
  \city{Charlottesville}
  \state{Virginia}
  \country{USA}
}
\email{lu.feng@virginia.edu}

\author{Ufuk Topcu}
\affiliation{
  \institution{University of Texas at Austin}
  \city{Austin}
  \state{Texas}
  \country{USA}
}
\email{utopcu@utexas.edu}


\begin{abstract}
    Large Language Models (LLMs) are increasingly used for planning tasks, offering unique capabilities not found in classical planners such as generating explanations and iterative refinement. However, trust---a critical factor in the adoption of planning systems---remains underexplored in the context of LLM-based planning tasks. This study bridges this gap by comparing human trust in LLM-based planners with classical planners through a user study in a Planning Domain Definition Language (PDDL) domain. Combining subjective measures, such as trust questionnaires, with objective metrics like evaluation accuracy, our findings reveal that correctness is the primary driver of trust and performance. Explanations provided by the LLM improved evaluation accuracy but had limited impact on trust, while plan refinement showed potential for increasing trust without significantly enhancing evaluation accuracy.
\end{abstract}







\maketitle

\section{Introduction} \label{sec:intro} 
Planning is the process of determining a sequence of actions to transit from an initial state to a desired goal state. Planners---systems designed to generate such action sequences under given constraints---play a critical role in automating decision-making processes in domains such as robotic navigation, logistics optimization, and medical scheduling.

Traditional planners, while effective in structured and predictable environments, often struggle with rigidity and lack of explainability. In contrast, Large Language Models (LLMs) have recently demonstrated strong performance in various domains, including text generation~\cite{li2024pre}, question answering~\cite{puri2020training,ram2021few}, and code completion~\cite{liu2020multi}. Unlike traditional planners, 
LLMs support multi-plan generation (i.e., return multiple plans to enable users to choose), dynamic adjustments based on externally given information, and understandable communication with humans via natural language.
These strengths have sparked growing interest in using LLMs as planners across diverse domains, including robotics~\cite{ren2023robots,singh2023progprompt,verifiable-SDM,huang2022language}, healthcare~\cite{cascella2023evaluating,sallam2023chatgpt}, and law~\cite{wu2023precedent,cheong2024not}.

However, the increasing use of LLM-based planners raises concerns, particularly regarding trust. Trust, defined as the willingness to rely on automated systems~\cite{lee2004trust}, is vital for the adoption of planning systems. Without trust, even systems with superior technical capabilities may struggle to gain acceptance in practical settings~\cite{vorm2022integrating}. 
Planning tasks are uniquely challenging due to their reliance on high correctness, sequential reasoning, and the need for robust adaptation to dynamic environments~\cite{allmendinger2017planning}.
These factors amplify the importance of trust, as both over-trust and under-trust can introduce errors or inefficiencies in planning and can have cascading effects on task success~\cite{talvitie2012problem,laurian2009trust}. Thus, fostering appropriate trust levels in LLM-based planners is essential for maximizing their potential while minimizing risks.

While prior research has explored factors influencing trust in LLM-based systems, such as anthropomorphic cues~\cite{cohn2024believing}, the framing and presence of explanations~\cite{sharma2024would}, and user interface design~\cite{sun2024trust}, factors influencing human trust in LLMs in the context of planning tasks remain underexplored. 
As the Planning Domain Definition Language (PDDL) has become a common benchmark for evaluating the planning capabilities of LLMs~\cite{silver2022pddl, silver2024generalized}, existing work primarily focuses on technical performance metrics, such as plan correctness and efficiency. 
To the best of our knowledge, no prior studies have empirically investigated human trust in LLM-based planners compared to classical PDDL solvers in a PDDL domain.
\emph{This work bridges this gap by conducting an exploratory user study that evaluates trust in a PDDL domain}.

Specifically, LLMs possess unique capabilities and limitations compared to classical PDDL planners~\cite{mcdermott20001998, ghallab2004automated} that may affect trust levels.
For instance, LLMs can generate natural language explanations to clarify why specific decisions were made~\cite{DBLP:conf/nips/WangCCLML23,dobhal2024using} and iteratively refine their outputs based on user feedback~\cite{RLHF,Christiano2017RLHF,Ouyang2022FollowInstructions,DBLP:conf/mlsys/YangBIWCWT24}. These capabilities have been shown in other contexts to enhance user trust by making the planning process more transparent and interactive~\cite{kunkel2019let, sebo2019don}. However, LLMs also exhibit significant limitations, such as their inability to reliably generate or validate plans independently, even for relatively simple tasks~\cite{kambhampati2024llms, valmeekam2022large, silver2022pddl, valmeekam2023planning}.
These capabilities and limitations highlight the need for a deeper understanding of the interplay among correctness, explanation, and refinement.

Trust can be evaluated using Likert-scale user questionnaires~\cite{martelaro2016tell,xu2015optimo,choi2015investigating} and broader instruments like the Propensity to Trust scale~\cite{merritt2013trust}, which assesses general attitudes toward machines. This study combines 7-point Likert scale trust scores as a subjective metric with users' evaluation accuracy of generated plans as an objective metric.

Key findings of our study reveal correctness as the dominant factor influencing both evaluation accuracy and trust, with the PDDL solver achieving the highest scores in both metrics. While explanations provided by the LLM planner enhanced evaluation accuracy, they had minimal effect on trust. Conversely, plan refinement showed potential to increase trust without improving evaluation accuracy. Notably, the results on refinement suggest that LLMs can gain user trust without genuine improvements in their planning capabilities, as the refined plan is generated using the same underlying model. This phenomenon underscores a critical implication: as most LLMs are fine-tuned via subjective human feedback~\cite{RLHF, Christiano2017RLHF}, the fine-tuned models tend to generate outputs that comply with human preferences, i.e., humans perceive as appealing or trustworthy, instead of objective correctness. Such tendencies could be exploited, intentionally or inadvertently, to potentially foster overtrust. Overall, the study results offer practical insights for designing human-centered AI planning systems.


\section{Methods} \label{sec:method} 
We evaluate factors influencing user trust in planners by comparing a language-model-based planner, denoted as an \emph{LLM Planner} (GPT-4o~\cite{achiam2023gpt}), with a traditional graph-search-based planner, denoted as a \emph{PDDL Solver} (Fast Downwards~\cite{fastdownward}). Unlike the PDDL Solver which relies on graph search algorithms, the LLM Planner can reason through the planning problem, explain its proposed solution, and iteratively refine the solution based on external feedback.

\subsection{Planning Problem}
A planning problem consists of a \emph{planning domain} (aspects of a problem that remains consistent, i.e. objects, predicates, actions) and a \emph{problem description} (particular instance of a planning task, i.e. initial state, goal state), expressed in PDDL. We present an example of the gripper problem in \Cref{app: grippers}.

We select the \emph{gripper} planning problems from the International Planning Competition~\cite{IPC} for plan generation and evaluation. In a gripper planning problem, a robot moves balls between a set of rooms using two grippers. The objective is to create a \emph{plan}---a sequence of actions---for the robot to move the balls to the defined target rooms. We present a few running examples of the gripper problem in \Cref{app: llm-planner} (\Cref{fig: correctness}).

\begin{figure}[t]
    \centering
    \includegraphics[width=0.75\linewidth]{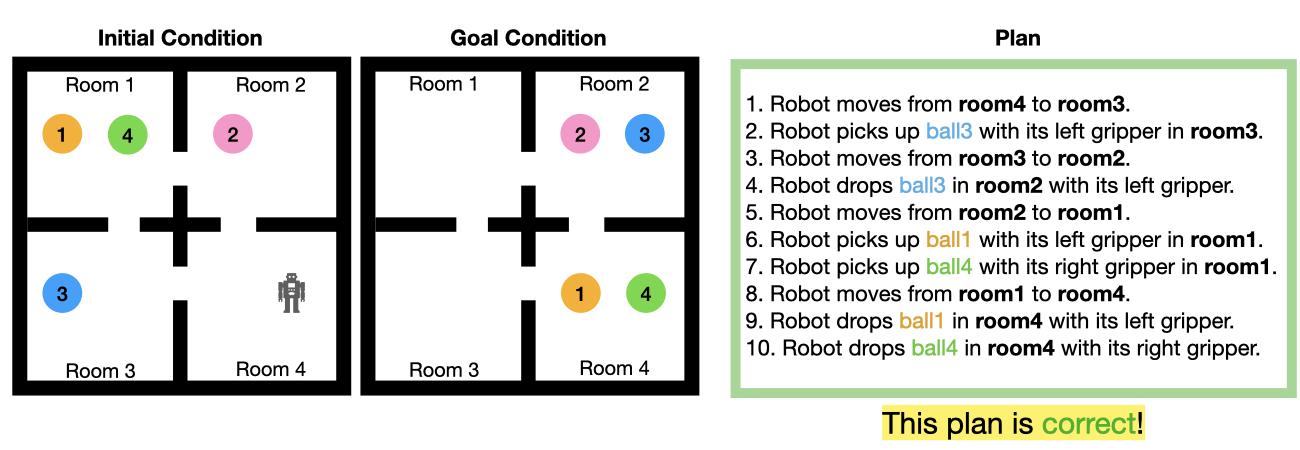}
    \caption{Examples where LLM Planner correctly generates a plan for the gripper planning problem.}
    \Description{Planning Problem Correctness}
    \label{fig: correct}
\end{figure}

\subsection{PDDL Solver}
The PDDL Solver takes the planning domain and the problem description as inputs and then generates a plan (a sequence of actions with specific input parameters) described in PDDL. 
Next, we convert the generated plan into natural language for user studies following the procedure in~\cite{seipp-et-al-zenodo2022} and display it to users. We present an example in Figure \ref{fig: correct}.
The planner either generates a \emph{correct} plan defined as the shortest path between the initial and goal states or returns a signal indicating that no solution exists for the given problem.

\subsection{LLM Planner}
The LLM Planner addresses planning problems by querying a large language model, using a structured prompt format. The planner then retrieves a natural language plan from the language model. To ensure the output adheres to the desired format, we include a few in-context examples within the prompts. We present an example of this in Appendix \ref{app: llm-planner}.
Unlike the PDDL Solver, the LLM Planner may generate \emph{incorrect} plans that violate the problem specifications (e.g., preconditions of actions) or fail to achieve the goal, as language models may struggle with large state spaces compared to classical planners.

\paragraph{LLM Planner with Explanation (LLM+Expl)}
To examine the influence of explanation on user trust, we create a natural language explanation of each generated plan. The trust improvement by adding explanations will motivate training an LLM to explain its plan.
This explanation includes an assessment of the plan’s correctness, identification of any violations of action preconditions, and an analysis of inconsistencies between the final state achieved and the intended goal state.
If a plan is correct, the explanation is ``the plan successfully satisfies the goal conditions.'' 
If a plan is incorrect, we identify the underlying cause as either a violation of action preconditions or a failure to achieve the goal state. In cases involving precondition violations, we specify the action responsible for the issue. 

For example, consider the action ``robot moves from room 1 to room 2,'' but the robot is initially located in room 3. This scenario constitutes a violation of the precondition for the ``move'' action. In the latter case, we describe the differences between the final state achieved and the intended goal state, e.g., ``fail to move ball 2 to room 2.''
This function enables the user to better understand why actions are chosen and their effect on the overall plan.
We present examples of explanations in \Cref{app: llm-planner} (\Cref{fig: explain}).

\paragraph{LLM Planner with Refinement (LLM+Refine)}
Refining an LLM-generated plan is also possible. So, we offer a prompting mechanism for the LLM Planner to refine the generated plan according to the user feedback. We present a sample user interface on the left of \Cref{fig: refine} in \Cref{app: llm-planner}. The mechanism works as follows:
First, request the user to indicate the step number where refinement should begin.
Second, send the planning domain, problem description, and the original plan to the language model. Next, query the model to rewrite the subsequent steps starting from the user-specified step number.
Finally, replace the original plan with the newly refined plan and display it to the user.
This mechanism enables the user to focus on a subset of steps, facilitating a deeper interpretation of those actions. However, the correctness of the refined plan is still not guaranteed.

\section{User Study Design} \label{sec:study} 
We conducted a user study via Qualtrics to evaluate human trust in plans generated by the planners discussed above.~\footnote{This study was approved by IRB\#****. Study details are included in \Cref{appendix:user_study_details}.} footnote{This study was approved by IRB\#7035 and University of Texas at Austin IRB \#6873. Study details are included in \Cref{appendix:user_study_details}.} 

\subsection{Participants}
We recruited 30 participants through Prolific~\cite{palan2018prolific} (fluent English speakers over the age of 18). After informed consent, a reCAPTCHA test was administered as a bot check. To encourage engagement and accurate responses, participants were offered \emph{bonus payments} based on their evaluation accuracy, defined as correctly accepting correct plans and rejecting incorrect ones. 

The participants (80\% male, 17\% female, and 3\% preferred not to say) had an average age of $34.00$ (SD=$10.11$). Regarding prior experience with large language models (LLMs), $80\%$ of participants reported having used LLMs before, while $20\%$ had not. When asked about the frequency of using LLMs specifically for planning tasks, $33\%$ indicated that they use them frequently, $43\%$ occasionally, and $23\%$ never.

\begin{figure}
    \centering
    \includegraphics[width=0.9\linewidth]{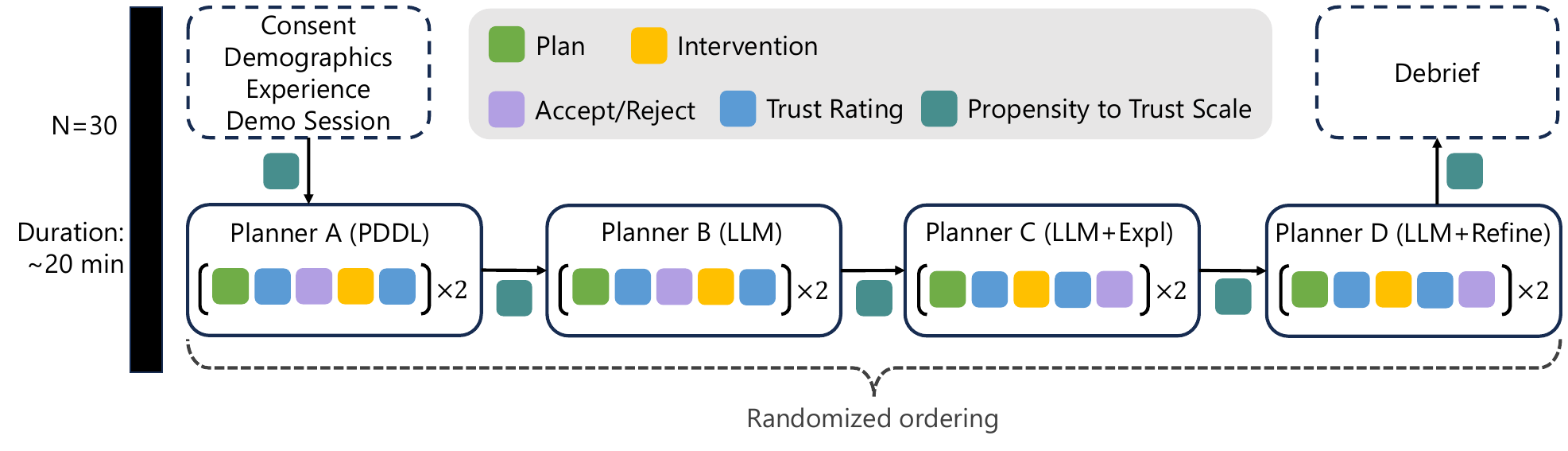}
    \caption{User study procedure.}
    \label{fig:procedure}
\end{figure}

\subsection{Procedure}
Participants began with a \emph{demo session} to familiarize themselves with the Gripper problem task and the study interface. 

The main part of the study comprised \emph{four sessions}, each corresponding to a different AI planner: \emph{Planner A (PDDL)}, \emph{Planner B (LLM)}, \emph{Planner C (LLM+Expl)}, and \emph{Planner D (LLM+Refine)}. The four sessions were presented in a randomized order to counterbalance any ordering effects. 

Each session contained \emph{two tasks}, each involving a new Gripper problem (unique initial and goal conditions) with similar difficulty (similar number of plan steps for equal number of rooms and balls). The order of tasks within each session was also randomized.
In each task, participants were first presented with a \emph{plan} generated by the planner and asked to rate their {trust} in the planner (\textbf{trust before}). Participants were then shown an \emph{intervention}, which varied depending on the session planner:
\begin{itemize}
    \item For \textbf{PDDL} and \textbf{LLM}, the intervention provided only the \textit{consequence of the plan}, e.g., \textit{``This plan is correct/wrong!''}.
    \item For \textbf{LLM+Expl}, the intervention included both the consequence of the plan and an \textit{explanation} of the outcome, e.g., \textit{``This plan is wrong because the robot misses the steps of moving ball4 from room4 to room1.''}
    \item For \textbf{LLM+Refine}, participants were first asked to \textit{choose between two lines} of the plan as a starting point for refinement. A \textit{revised plan} was then generated beginning from the selected line.
\end{itemize}
After the intervention, participants were again asked to rate their {trust} in the planner (\textbf{trust after}). 
Additionally, participants were asked to decide whether to \textbf{accept or reject} the plan before the intervention for PDDL and LLM planners but after for the other two. This allows evaluation of plan correctness prior to consequences for PDDL and LLM, while focusing on responses to interventions for the others.

At the end of the study, participants were informed of their \textit{evaluation accuracy} as the total number of correctly evaluated tasks out of 8 total tasks (2 tasks per session, 4 sessions in total).
The procedure is detailed in \Cref{fig:procedure}.

\subsection{Independent Variables}
We employed a within-subject design where each participant completes four sessions, each involving one of four planners. The PDDL solver provides 100\% correct plans, while the other three planners deliver 50\% correct plans. 
We set this accuracy to ensure non-perfect but meaningful performance across two tasks per session, approximating the observed accuracy in practice~\cite{zuo2024planetarium,hao2024planning}.
The LLM+Expl planner explains why the plans are correct or not, and the LLM+Refine planner allows participants to refine the plans.
The independent variables include correctness (comparing PDDL with LLM), explanation (comparing LLM with LLM+Expl), and refinement (comparing LLM with LLM+Refine).

\subsection{Dependent Measures}
For each session, participants evaluated two tasks. We measured user performance on \emph{evaluation accuracy} by the number of correctly evaluated tasks ($0$, $1$, or $2$). 
Participants also rated their \emph{trust} in the planner on a 7-point Likert scale (1 = strongly disagree, 7 = strongly agree) both \emph{before} and \emph{after} the intervention.
We also measured participants' \emph{propensity to trust} at the end of each session using a six-item scale~\cite{merritt2013trust} to assess their general tendency to trust AI planners. The response options were on a 5-point Likert-type scale ranging from 1 (strongly disagree) to 5 (strongly agree). In our survey, we adapted the scale by replacing the term ``machine" with ``AI planner" to reflect the context of our study better (see \Cref{appendix:propensity_to_trust_scale} for the exact scale we used). 

\subsection{Hypotheses}
We formulated the following hypotheses to examine the effects of correctness, explanations, and refinement on user performance (plan evaluation accuracy) and trust:
\begin{itemize}
    \item \textbf{H1:} Planners that are more correct increase evaluation accuracy.
    \item \textbf{H2:} Planners that provide explanations increase evaluation accuracy.
    \item \textbf{H3:} Planners that allow for plan refinement increase evaluation accuracy. 
    \item \textbf{H4:} Planners that are more correct improve user trust.
    \item \textbf{H5:} Planners that provide explanations improve user trust.
    \item \textbf{H6:} Planners that allow for plan refinement improve user trust. 
\end{itemize}

\section{Results \& Analysis} \label{sec:results} 
This section presents findings from our user study on evaluation accuracy, user trust, and the propensity to trust scale.

\begin{figure}
    \centering
    \begin{minipage}[b]{0.72\textwidth} 
        \centering
        \begin{subfigure}[b]{0.32\textwidth}
            \centering
            \includegraphics[width=\linewidth]{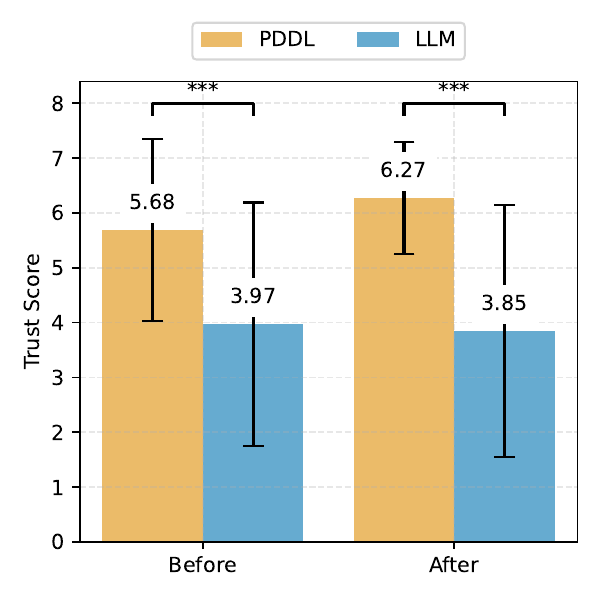}
            \caption{PDDL vs. LLM.}
            \label{fig:trust_bar_pddl_llm}
        \end{subfigure}%
        \hfill
        \begin{subfigure}[b]{0.32\textwidth}
            \centering
            \includegraphics[width=\linewidth]{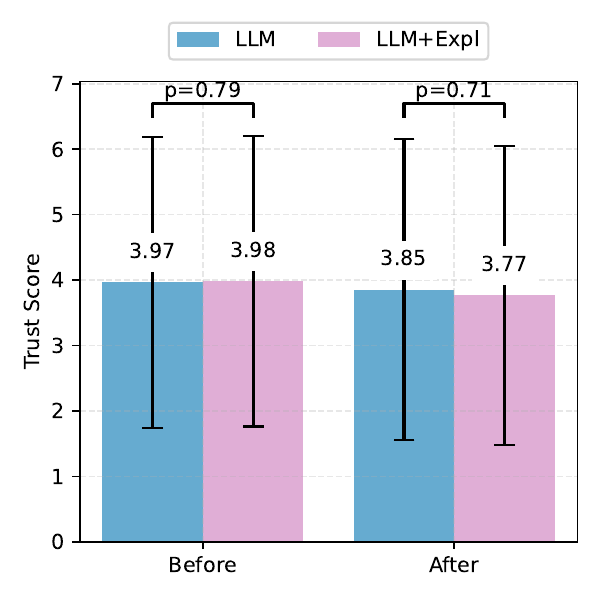}
            \caption{LLM vs. LLM+Expl.}
            \label{fig:trust_bar_expl}
        \end{subfigure}%
        \hfill
        \begin{subfigure}[b]{0.32\textwidth}
            \centering
            \includegraphics[width=\linewidth]{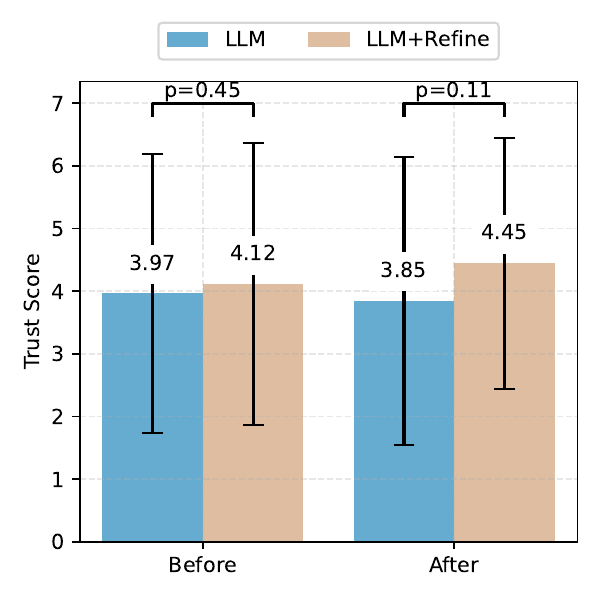}
            \caption{LLM vs. LLM+Refine.}
            \label{fig:trust_bar_repair}
        \end{subfigure}
        \caption{Trust scores on a 7-point Likert scale before and after.}
        \label{fig:trust_bars}
    \end{minipage}%
    \hfill
    \begin{minipage}[b]{0.23\textwidth}
        \centering
        \includegraphics[width=\linewidth]{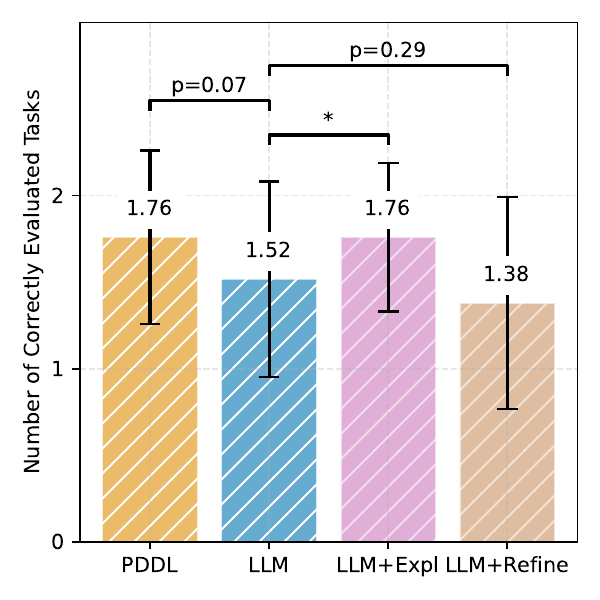}
        \caption{Evaluation accuracy measured by the number of correctly evaluated tasks.}
        \label{fig:correctness_bar}
    \end{minipage}
\end{figure}

\subsection{On Evaluation Accuracy}
\Cref{fig:correctness_bar} presents the average number of correctly evaluated tasks for each planner, with error bars indicating standard deviations. We used the Wilcoxon signed-rank test to evaluate our hypotheses H1-H3.

For \textbf{H1}, participants achieved an average accuracy of $1.76 \pm 0.50$ with the PDDL solver, compared to $1.52 \pm 0.56$ for the LLM planner. This result supports our hypothesis that correctness is a key determinant of evaluation accuracy. However, the difference was not statistically significant ($W=18, Z=-4.31, p=0.071, r=-0.801$). We suspect that increasing the sample size could reduce this uncertainty and strengthen the observed trend.

For \textbf{H2}, evaluation accuracy improved for the LLM planner when explanations were provided (LLM+Expl), reaching $1.76 \pm 0.43$. This difference was statistically significant ($W=5, Z=-4.59, p=0.020, r=-0.853$), supporting our hypothesis that planners with explanations increase evaluation accuracy.

For \textbf{H3}, the results for the LLM planner with refinement (LLM+Refine) did not align with our hypothesis. 
Participants achieved an average accuracy of $1.38 \pm 0.61$ with the LLM+Refine planner, compared to $1.52 \pm 0.56$ for the basic LLM planner. Although this result deviates from our hypothesis that refinement would improve evaluation accuracy, the difference is not statistically significant ($W=22, Z=-4.23, p=0.285, r=-0.785$). As a result, we cannot conclusively confirm or refute H3. One possible explanation for this deviation is overtrust: Participants may assume that the opportunity to revise the plan ensures the planner would correct itself, leading them to evaluate the revised plan less critically and, consequently, with lower accuracy.

\emph{Thus, the data suggests support for H1, confirms H2, and suggests rejection of H3.}

\subsection{On Trust}
\Cref{fig:trust_bars} shows participants' average self-reported trust levels before and after each intervention, measured on a 7-point Likert scale, with error bars representing standard deviations. We used the Wilcoxon signed-rank test to evaluate our hypotheses H4-H6.

For \textbf{H4}, \Cref{fig:trust_bar_pddl_llm} shows that PDDL achieved statistically significantly higher trust levels than LLM both \textbf{before} the intervention ($W=134.5, Z=-5.75, p<0.001, r=-0.742$) and \textbf{after} ($W=19, Z=-6.60, p<0.001, r=-0.852$). In terms of trust dynamics, participants' trust in PDDL significantly increased from $5.68 \pm 1.66$ to $6.27 \pm 1.02$ ($W=10, Z=-6.66, p=0.001, r=-0.860$). In contrast, trust in LLM showed a slight decrease from $3.97 \pm 2.22$ to $3.85 \pm 2.30$, though this change was not statistically significant ($W=215.50, Z=-5.15, p=0.722, r=-0.665$). These findings support the hypothesis that the correctness of planners is a key factor influencing human trust.

For \textbf{H5}, \Cref{fig:trust_bar_expl} shows no statistically significant difference in trust levels between LLM and LLM+Expl, both before and after the intervention. This result challenges our hypothesis that providing explanations would increase trust when correctness is controlled. 
One possible interpretation is that participants primarily value the objective correctness of the plans, with explanations offering little benefit unless correctness improves.
Alternatively, explanations may help participants calibrate their trust by revealing the planner's limitations, allowing them to adjust their trust to appropriate levels. This insight suggests that improving trust in LLMs for planning tasks may require prioritizing the objective correctness of the plans over supplementary explanations.

For \textbf{H6}, \Cref{fig:trust_bar_repair} shows a slight increase in trust levels with LLM+Refine. On average, trust rose from $3.97 \pm 2.22$ to $4.12 \pm 2.25$ before the intervention and from $3.85 \pm 2.30$ to $4.45 \pm 2.00$ after. While this trend is not statistically significant, it suggests a potential positive effect of refinement on human trust with the LLM planner.

\emph{Thus, the data supports H4, suggests rejection of H5, and suggests support of H6.}

\begin{figure}
    \centering
    \includegraphics[width=\linewidth]{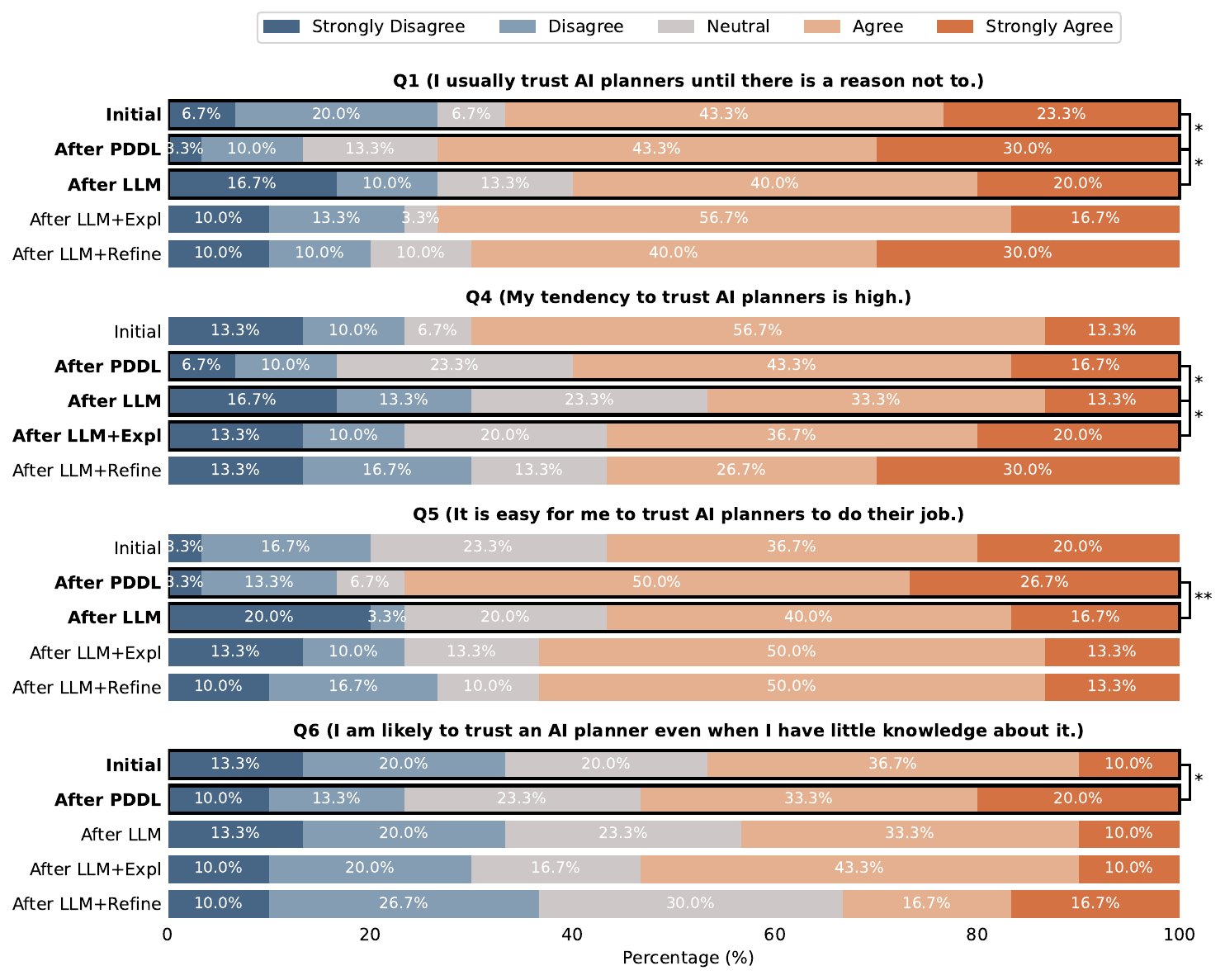}
    \caption{Propensity to Trust Scale: 4 out of 6 Questions with Statistically Significant Differences Highlighted (*$p<0.05$, **$p<0.01$).}
    \label{fig:propensity_to_trust_scale}
\end{figure}
\subsection{Propensity to Trust Scale}
The \emph{propensity to trust} scale consists of six questions used to assess an individual's tendency to trust machines based on their current behaviors~\cite{merritt2013trust}.
While we did not have specific hypotheses tied to this scale, we included it to gather initial insights for future exploration. 
We found that four questions showed statistically significant differences between planners using the Wilcoxon signed-rank test and are displayed in \Cref{fig:propensity_to_trust_scale}. Full results are provided in the appendix.

For \textbf{Q1} and \textbf{Q6}, we observe a clear shift toward agreement after the PDDL condition compared to the initial baseline. This suggests that participants were more inclined to trust AI planners following the PDDL session, likely due to the 100\% correctness of PDDL plans, which appears to boost trust.
In contrast, for \textbf{Q1}, \textbf{Q4}, and \textbf{Q5}, we see a notable reduction in agreement after interacting with the LLM planner compared to the PDDL solver. This decrease aligns with the reduced correctness of the LLM plans (50\%), highlighting the importance of correctness in maintaining trust in AI planners.
Interestingly, \textbf{Q4} reveals that providing explanations (LLM+Expl) helps recover participants' agreement levels compared to the basic LLM condition. However, this positive effect of explanations on trust propensity is limited, as it is only observed in one of the six questions.

These results underscore that correctness remains the dominant factor influencing participants' general trust attitude towards AI planners, with explanations offering only minimal benefit when correctness is suboptimal.

\section{Discussion} \label{sec:conclu} 
\paragraph{Summary}
Our findings provide significant insights into the influence of correctness, explanations, and refinement on evaluation accuracy and user trust in AI-based planners. 
In particular, the findings are three-fold: 
(1) The \textbf{correctness} of the generated plans is the most significant factor that impacts the evaluation accuracy and user trust in the planners. As the PDDL solver is more capable of generating correct plans, it achieves the highest evaluation accuracy and trust. 
(2) The \textbf{explanation} component of the LLM planner improves evaluation accuracy, as LLM+Expl achieves higher accuracy than LLM alone. Despite this improvement, LLM+Expl minimally impacts user trust. However, alternative explanation methods may influence user trust differently from the manually generated explanations used in our approach.
(3) The \textbf{refinement} procedure in the LLM planner does not lead to a significant improvement in evaluation accuracy; however, it exhibits a positive influence on user trust that may indicate an overtrust in some situations.
Finally, the propensity-to-trust analysis identifies correctness as the primary determinant of user trust, whereas explanations provided limited improvement in scenarios where the planner's accuracy is diminished.


\paragraph{Future Research} Future steps in this research include expanding user studies with larger sample sizes to improve generalizability and including additional planning problems per session for a more comprehensive evaluation. Next, we will explore alternative methods for generating plan explanations beyond manual creation to identify approaches that more effectively enhance user trust. 
Additionally, we will examine user trust by employing multiple LLM-based planners with varying levels of planning accuracy to better understand the interplay between planning correctness and user trust. 
Furthermore, we aim to enable real-time user-planner interaction, allowing users to provide feedback and refine plans collaboratively, thereby fostering a more dynamic and user-centric planning process.


\bibliographystyle{ACM-Reference-Format}
\bibliography{references}

\appendix
\newpage

\section{Gripper Planning Problem}
\label{app: grippers}
The types and predicates collaboratively define the states of a planning environment. Then, the actions define the transition of the environment. Each action consists of a set of input parameters, a precondition, and an effect. We consider the precondition and effect as the initial and final state of the action. 

The types of objects of the gripper problem are defined as
\vspace{4pt}
\begin{lstlisting}[language=completion]
(:types room ball robot gripper) 
<completion>; there are several balls distributed in several rooms and a robot with two grippers.</completion>
\end{lstlisting}
and the predicates are defined as
\vspace{4pt}
\begin{lstlisting}[language=completion]
(:predicates (at-robby ?r - robot ?x - room) <completion>; a predicate indicating the robot's location</completion>
    (at ?o - ball ?x - room) <completion>; a predicate indicating the ball's location</completion>
    (free ?r - robot ?g - gripper) <completion>; a predicate indicating whether the robot's gripper is free</completion>
    (carry ?r - robot ?o - ball ?g - gripper)) <completion>; indicating the ball carried by a gripper</completion>
\end{lstlisting}

An action moving the robot from a room to another room is defined as
\vspace{4pt}
\begin{lstlisting}[language=completion]
(:action move
    :parameters  (?r - robot ?from ?to - room) <completion>; we specify the initial and target rooms</completion>
    :precondition (and  (at-robby ?r ?from)) <completion>; the robot has to be in the initial room</completion>
    :effect (and  (at-robby ?r ?to) (not (at-robby ?r ?from))))
\end{lstlisting}
Furthermore, we have actions ``pick (a ball with a gripper)'' and ``drop (a ball).''
\vspace{4pt}
\begin{lstlisting}[language=completion]
(:action pick
       :parameters (?r - robot ?obj - object ?room - room ?g - gripper)
       :precondition  (and  (at ?obj ?room) (at-robby ?r ?room) (free ?r ?g))
       :effect (and (carry ?r ?obj ?g)
		    (not (at ?obj ?room)) 
		    (not (free ?r ?g))))

(:action drop
    :parameters (?r - robot ?obj - object ?room - room ?g - gripper)
    :precondition  (and  (carry ?r ?obj ?g) (at-robby ?r ?room))
    :effect (and (at ?obj ?room)
		(free ?r ?g)
		(not (carry ?r ?obj ?g)))))
\end{lstlisting}

An example of the initial and goal states are
\vspace{4pt}
\begin{lstlisting}[language=completion]
(:init (at-robby robot1 room1) (free robot1 rgripper1) (free robot1 lgripper1)
    (at ball1 room1) (at ball2 room3) (at ball3 room1) (at ball4 room2) )
(:goal (and (at ball1 room1) (at ball2 room3) (at ball3 room1) (at ball4 room2) ) )
\end{lstlisting}

\section{Additional Details on LLM Planner}
\label{app: llm-planner}

Figure \ref{fig: in-context}, \ref{fig: prompt}, and \ref{fig: refine-prompt} show the complete prompt for querying the language model to solve a planning problem. The blue text represents the prompts to the language model, while the red text corresponds to the responses generated by the language model.

\begin{figure}[H]
    \centering
    \begin{lstlisting}[language=completion]
    <prompt>User: Given the following planning domain:
    (define (domain gripper-strips)
       (:requirements :strips :typing) 
       (:types room object robot gripper)
       (:predicates (at-robby ?r - robot ?x - room)
                  (at ?o - object ?x - room)
                  (free ?r - robot ?g - gripper)
                  (carry ?r - robot ?o - object ?g - gripper))
       (:action move
           :parameters  (?r - robot ?from ?to - room)
           :precondition (and  (at-robby ?r ?from))
           :effect (and  (at-robby ?r ?to) (not (at-robby ?r ?from))))
       (:action pick
           :parameters (?r - robot ?obj - object ?room - room ?g - gripper)
           :precondition  (and  (at ?obj ?room) (at-robby ?r ?room) (free ?r ?g))
           :effect (and (carry ?r ?obj ?g) (not (at ?obj ?room)) (not (free ?r ?g))))
       (:action drop
           :parameters (?r - robot ?obj - object ?room - room ?g - gripper)
           :precondition  (and  (carry ?r ?obj ?g) (at-robby ?r ?room))
           :effect (and (at ?obj ?room) (free ?r ?g) (not (carry ?r ?obj ?g)))))
           
    Please solve the following PDDL problem and return the solution only.
    (define (problem gripper-1-2-1)
        (:domain gripper-strips)
        (:objects robot1 - robot
                  rgripper1 lgripper1 - gripper
                  room1 room2 - room
                  ball1 - object)
        (:init (at-robby robot2 room1)
               (free robot1 rgripper1)
               (free robot1 lgripper1)
               (at ball1 room1) )
        (:goal (at ball1 room2) ) )
    
    Assistant:
    1. Robot1 moves from room2 to room1. 
    2. Robot1 picks up ball1 with its left gripper in room1.
    3. Robot1 moves from room1 to room2. 
    4. Robot1 drops ball1 in room2 with its left gripper. </prompt>
    \end{lstlisting}
    \Description{In-context example.}
    \caption{An in-context example provided to the language model to constrain its output format.}
    \label{fig: in-context}
\end{figure}

\begin{figure}[H]
    \centering
    \begin{lstlisting}[language=completion]
    <prompt><In-context example from the previous textbox>
    
    User: Given the same planning domain:
    Please solve the following PDDL problem and return the solution only.
    (define (problem gripper-1-4-4)
        (:domain gripper-strips)
        (:objects robot1 - robot
                  rgripper1 lgripper1 - gripper
                  room1 room2 room3 room4 - room
                  ball1 ball2 ball3 ball4 - object)
        (:init (at-robby robot1 room1)
            (free robot1 rgripper1)
            (free robot1 lgripper1)
            (at ball1 room1)
            (at ball2 room3)
            (at ball3 room1)
            (at ball4 room2) )
        (:goal (and (at ball1 room1)
                (at ball2 room2)
                (at ball3 room2)
                (at ball4 room4) ) ) )
    Assistant: </prompt><completion>
    1. Robot1 picks up ball3 with its left gripper in room1.
    2. Robot1 moves from room1 to room2.
    3. Robot1 picks up ball4 with its right gripper in room2.
    4. Robot1 drops ball3 in room2 with its left gripper.
    5. Robot1 moves from room2 to room3.
    6. Robot1 picks up ball2 with its left gripper in room3.
    7. Robot1 moves from room3 to room2.
    8. Robot1 drops ball2 in room2 with its left gripper.
    9. Robot1 moves from room2 to room4.
    10. Robot1 drops ball4 in room4 with its right gripper.</completion>
    \end{lstlisting}
    \Description{Prompt}
    \caption{Complete prompt for querying a plan from the language model.}
    \label{fig: prompt}
\end{figure}

\begin{figure}[H]
    \centering
    \begin{lstlisting}[language=completion]
    <prompt>Given the planning domain and problem:
    (define (domain gripper-strips)
        ......)
    (define (problem gripper-1-2-1)
        (:domain gripper-strips)......)
    
    Please refine the plan from step 2:
    1. Robot1 moves from room2 to room1. 
    2. Robot1 picks up ball1 with its left gripper in room1.
    3. Robot1 picks up ball2 with its right gripper in room2.
    4. Robot1 drops ball3 in room2 with its left gripper. </prompt>
    <completion>1. Robot1 moves from room2 to room1. 
    2. Robot1 picks up ball1 with its left gripper in room1.
    3. Robot1 moves from room1 to room2. 
    4. Robot1 drops ball1 in room2 with its left gripper. </completion>
    \end{lstlisting}
    \Description{Refine prompt}
    \caption{Prompt for querying the LLM Planner to refine an existing plan.}
    \label{fig: refine-prompt}
\end{figure}

\begin{figure}[t]
    \centering
    \includegraphics[width=0.49\linewidth, trim=0 160 0 50, clip]{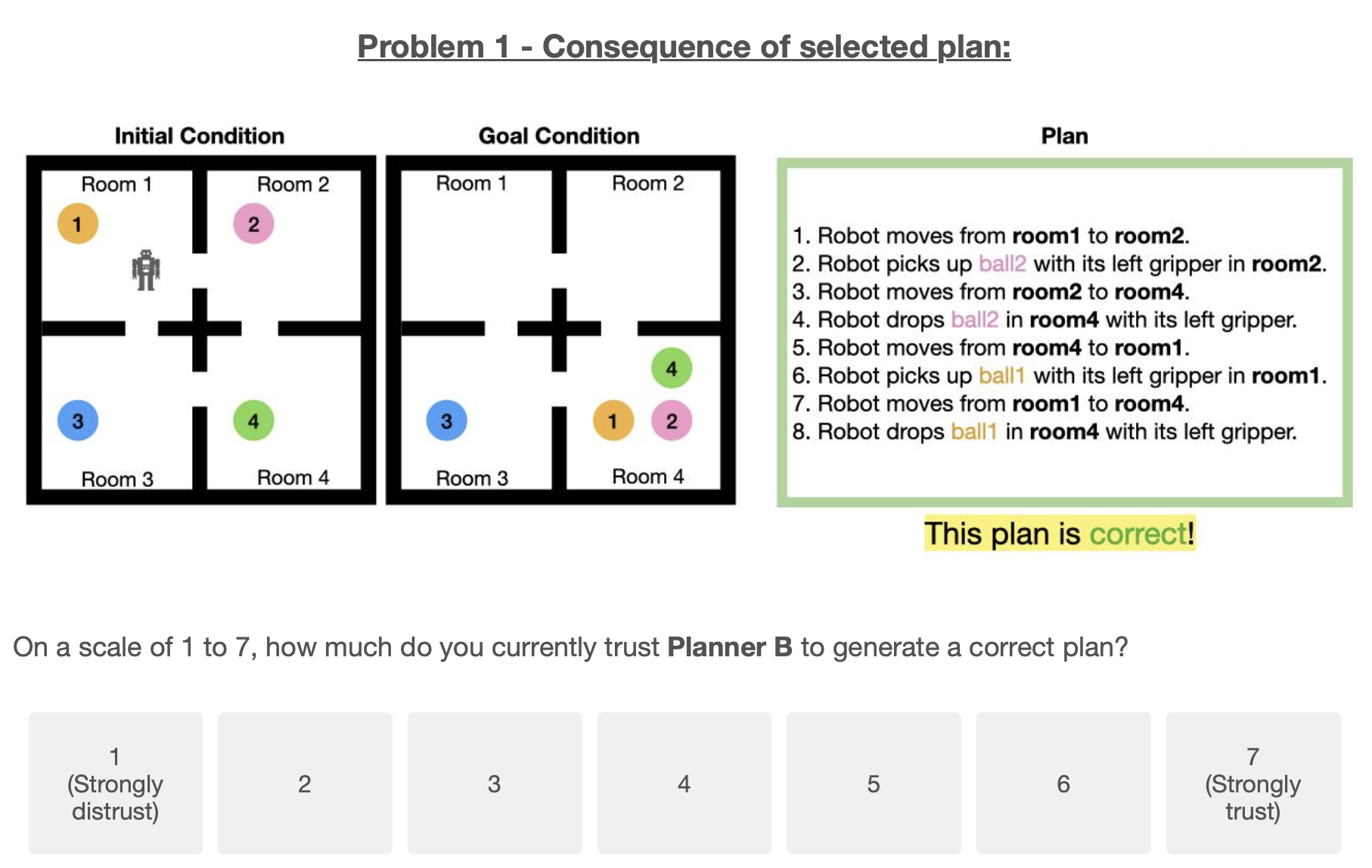}
    \includegraphics[width=0.49\linewidth, trim=0 170 0 50, clip]{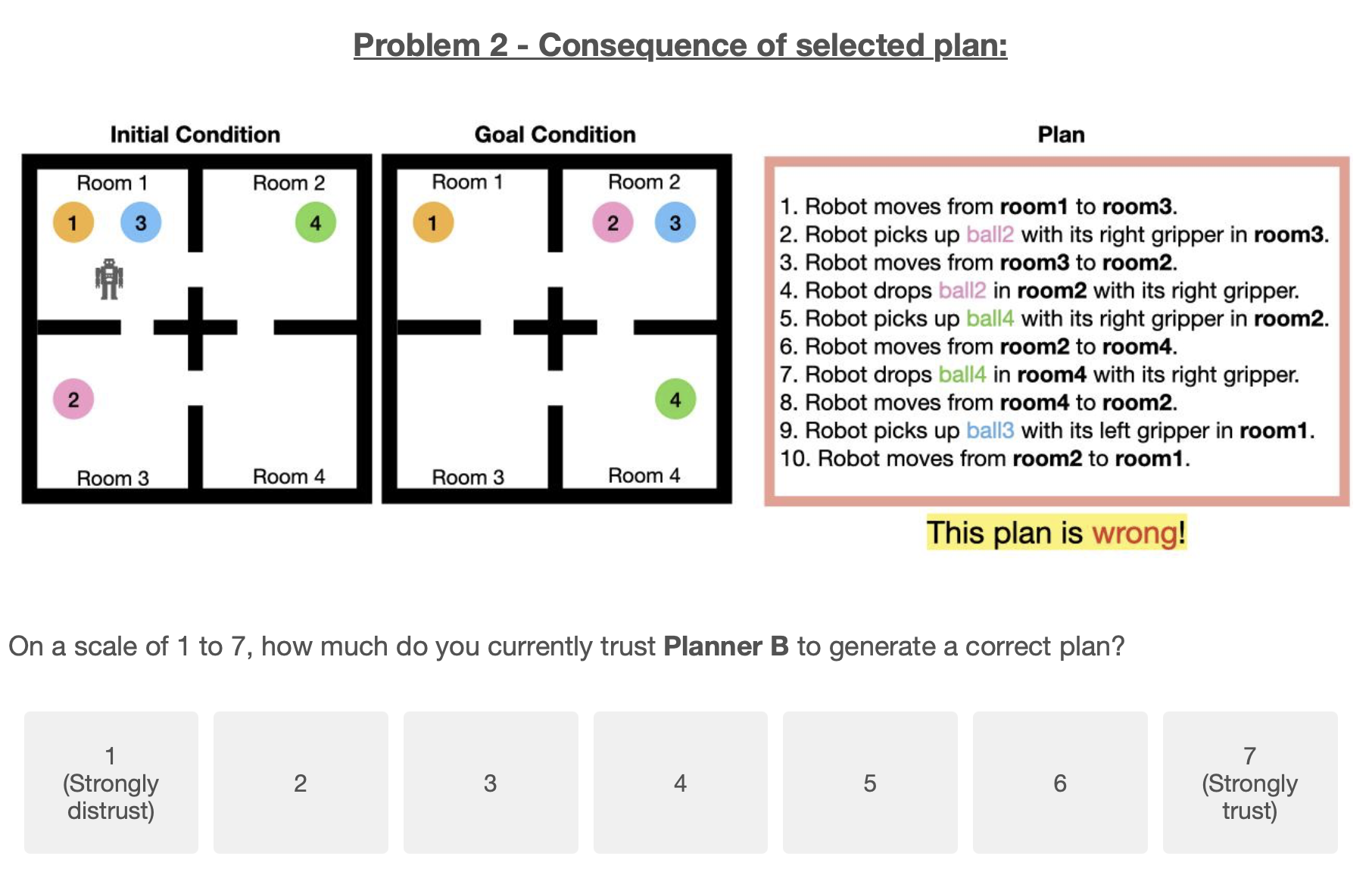}
    \caption{Examples where LLM Planner correctly and incorrectly generates plans for the gripper planning problems.}
    \Description{Planning Problem Correctness}
    \label{fig: correctness}
\end{figure}

\begin{figure}[t]
    \centering
    \includegraphics[width=0.49\linewidth, trim=0 350 0 0, clip]{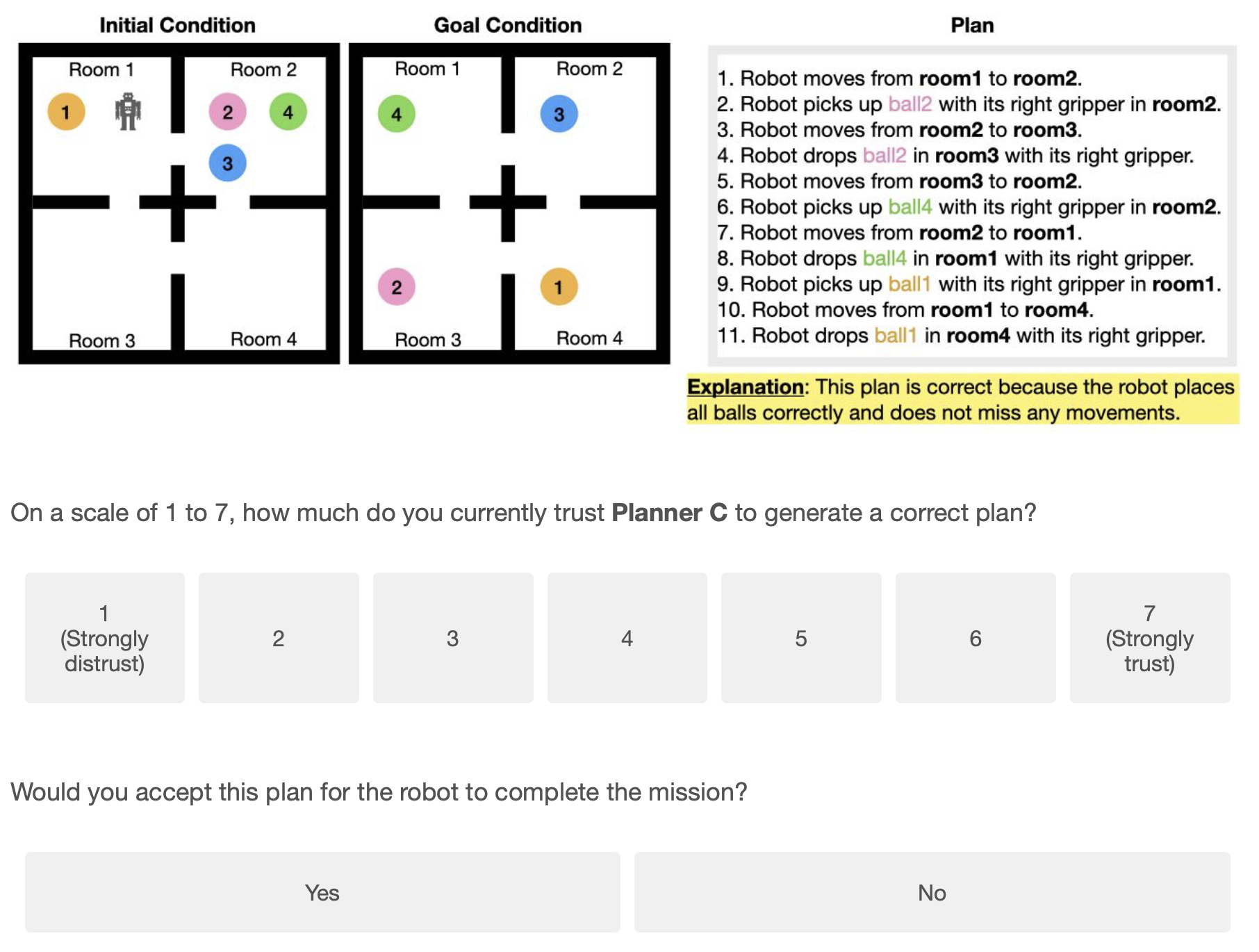}
    \includegraphics[width=0.49\linewidth, trim=0 350 0 0, clip]{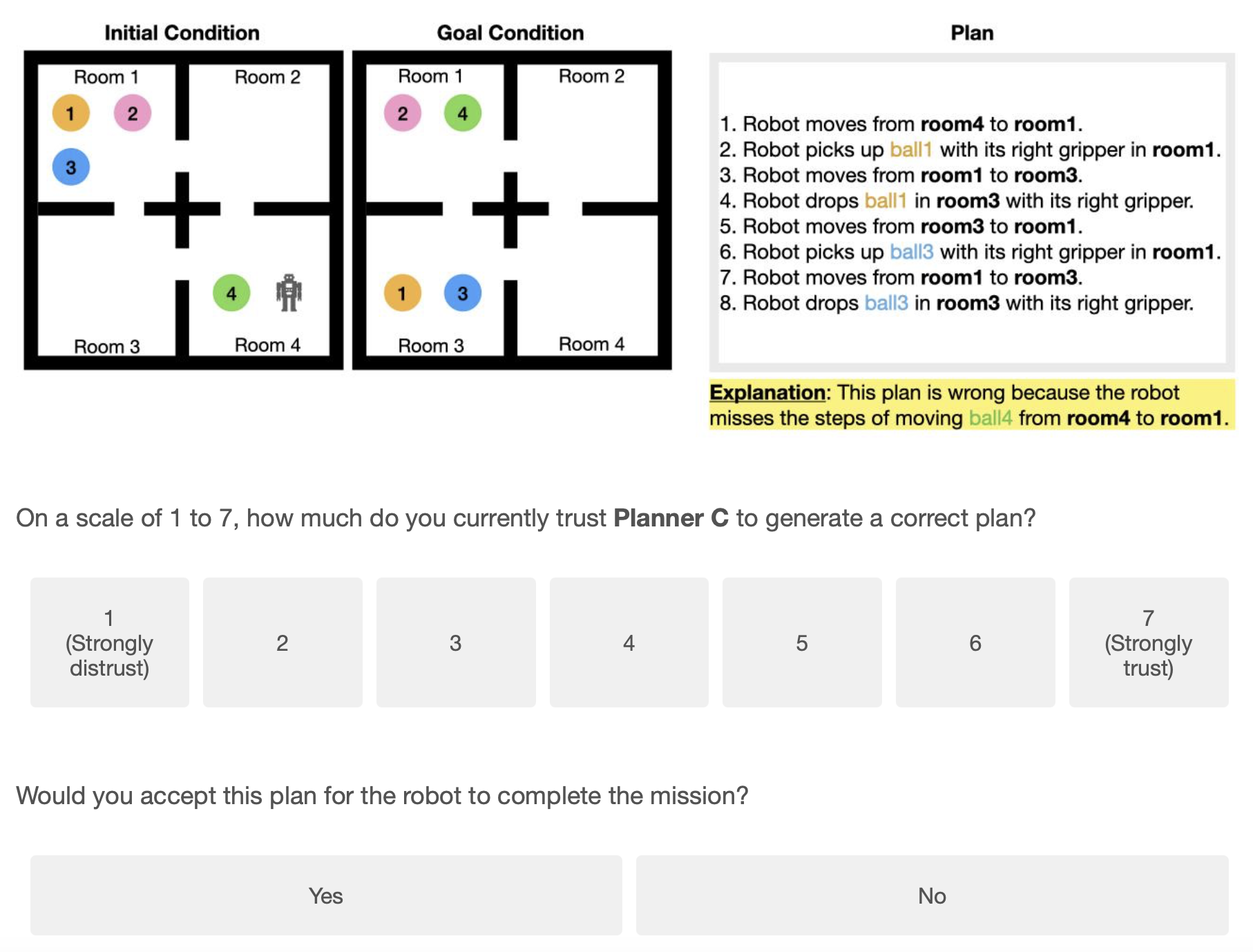}
    \caption{Explanations to the plans generated by the LLM Planner. The left example shows the explanation of a correctly generated plan while the right example shows the explanation of a incorrect plan.}
    \Description{Planning Problem Explanation}
    \label{fig: explain}
\end{figure}

\begin{figure}[t]
    \centering
    \includegraphics[height=0.24\linewidth]{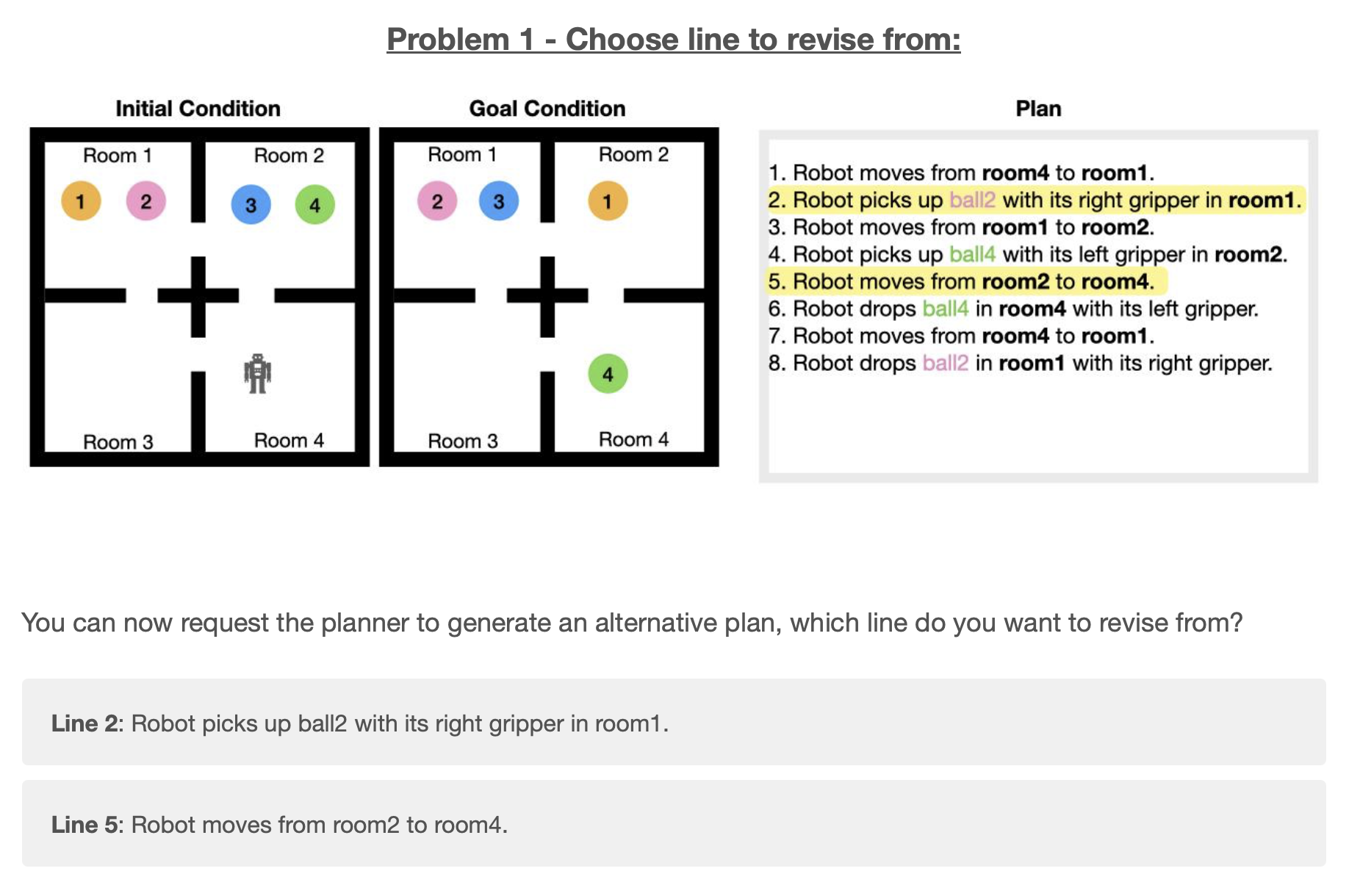}
    \includegraphics[height=0.24\linewidth, trim=500 405 10 0, clip]{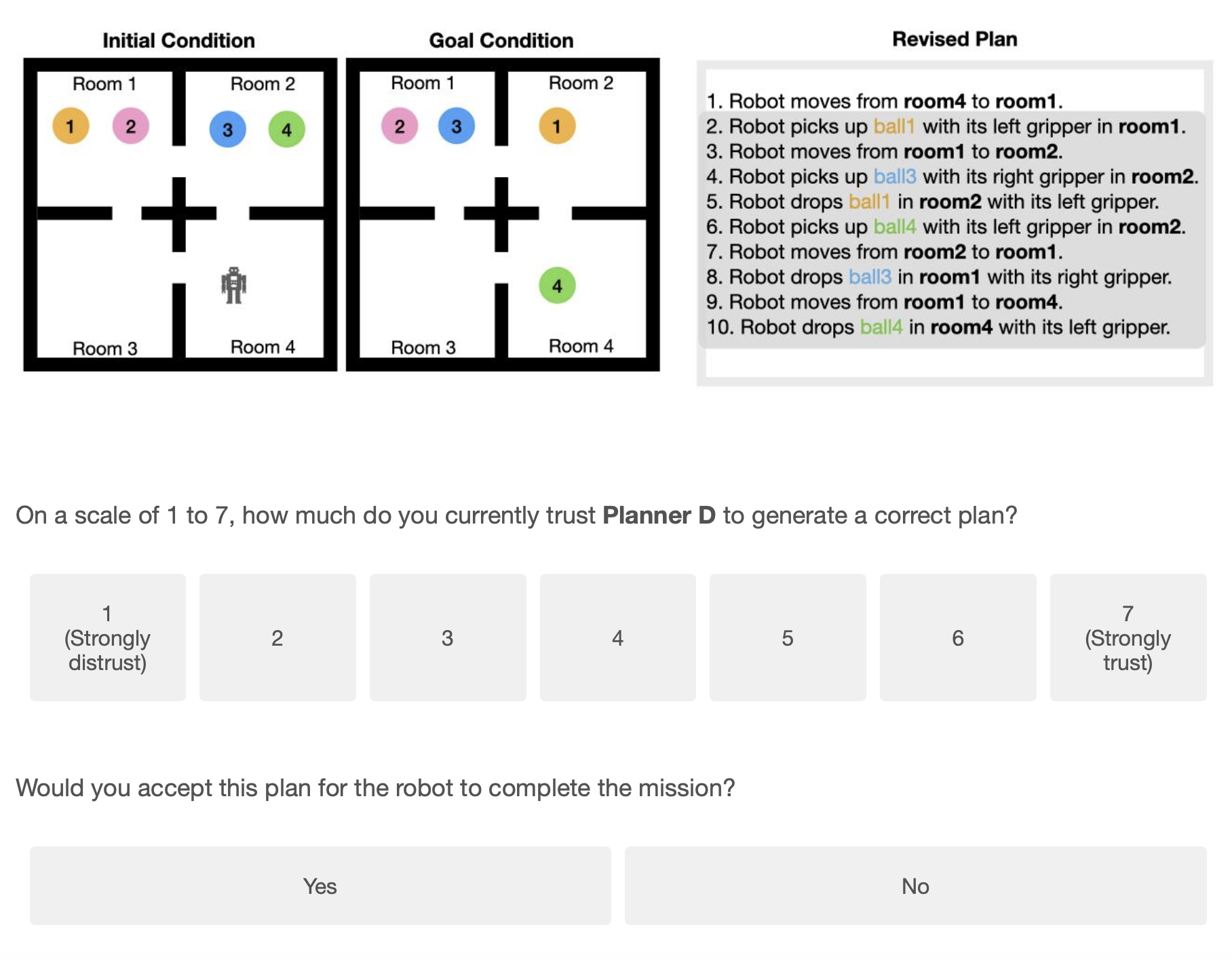}
    \includegraphics[height=0.24\linewidth, trim=500 400 20 0, clip]{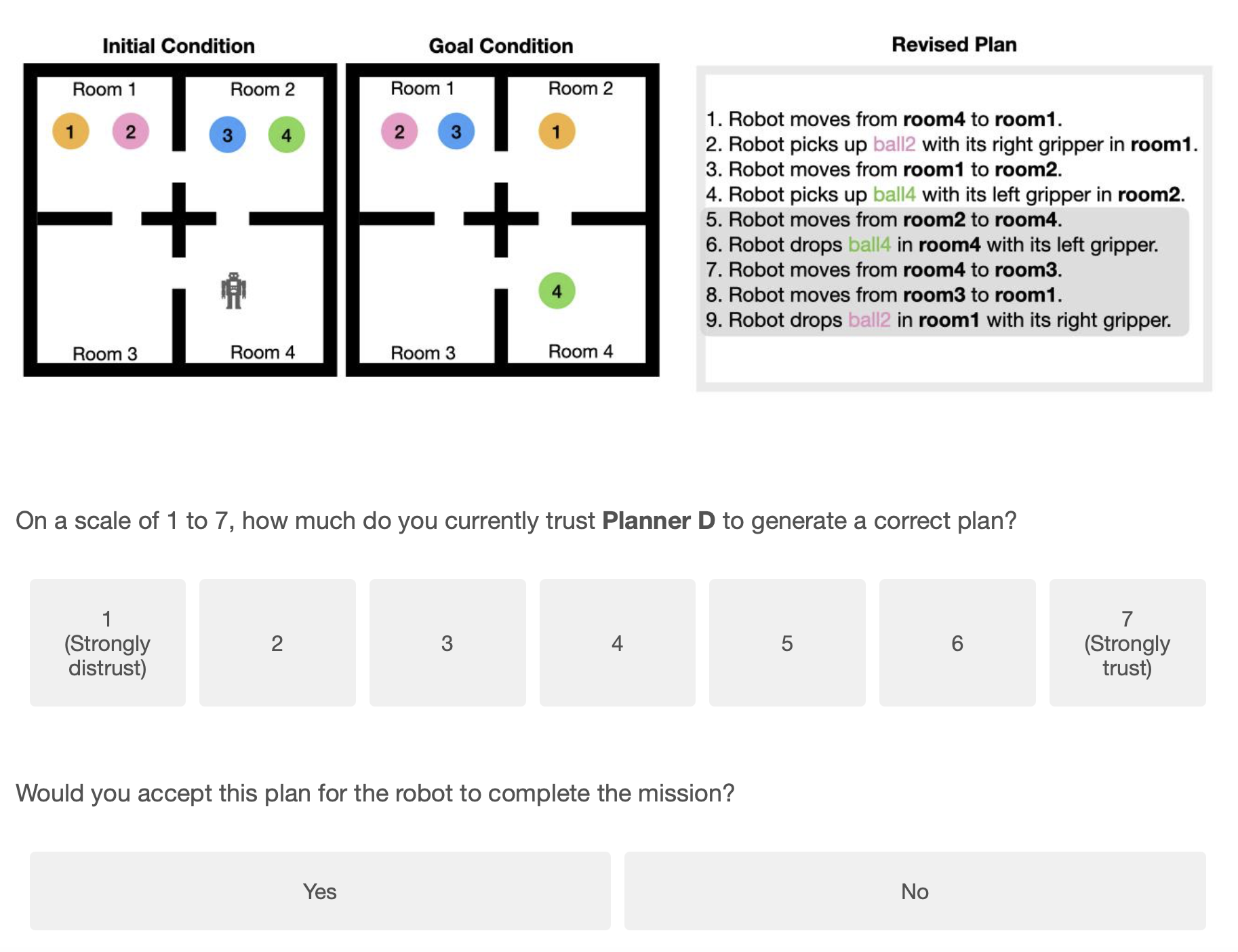}
    \caption{Plan refinement by the LLM Planner. The left figure shows the planning problem (initial and goal state). The text boxes from left to right present two choices of plan refinement (where the refinement starts), a correctly refined plan refined from step 2, and an incorrect plan refined from step 5.}
    \Description{Refinement}
    \label{fig: refine}
\end{figure}

\clearpage
\section{User Study Details}\label{appendix:user_study_details}

\subsection{Main Session Questionnaires}
\begin{figure}[ht]
    \centering
    \begin{minipage}[t]{0.48\linewidth}
        \centering
        \includegraphics[width=\linewidth]{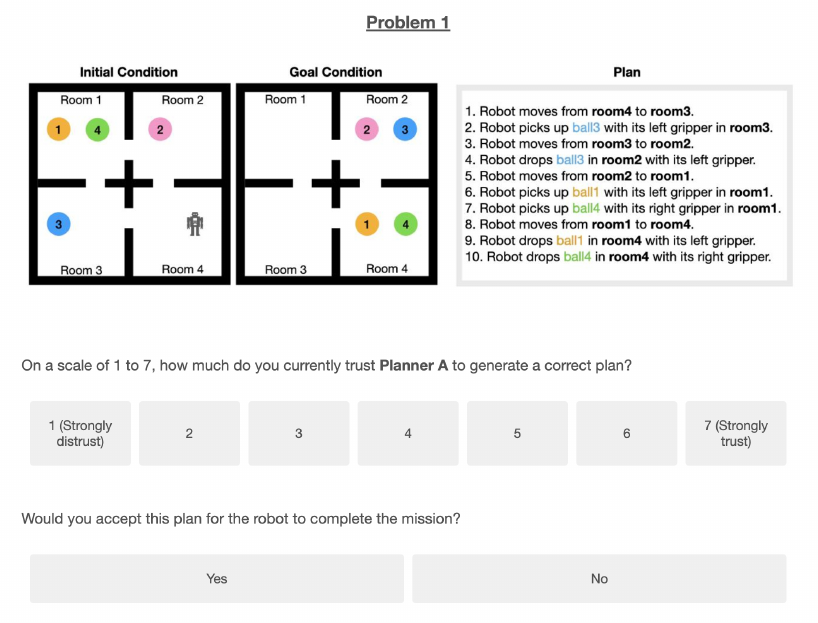}
        \caption{Planners A and B before intervention: Plan + Trust Rating + Accept/Reject.}
        \label{fig:before_intervention_ab}
    \end{minipage}
    \hfill 
    \begin{minipage}[t]{0.48\linewidth}
        \centering
        \includegraphics[width=\linewidth]{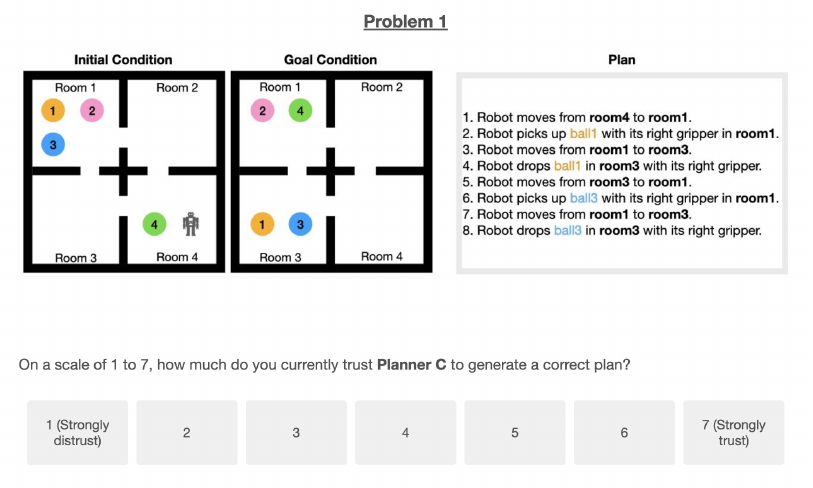}
        \caption{Planners C and D before intervention: Plan + Trust Rating.}
        \label{fig:before_intervention_cd}
    \end{minipage}
\end{figure}

\paragraph{Before Intervention.}
Participants were presented with different planners and asked to rate their trust in the plans and decide whether to accept or reject them. The setup differed slightly between planner groups, as shown in \Cref{fig:before_intervention_ab,fig:before_intervention_cd}.

\begin{figure}[ht]
    \centering
    \begin{minipage}[t]{0.48\linewidth}
        \centering
        \includegraphics[width=\linewidth]{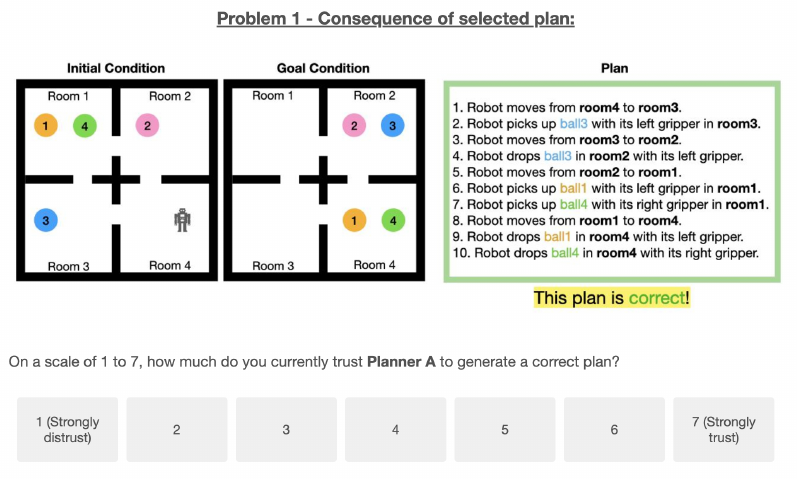}
    \label{fig:during_intervention_ab}
    \caption{Planners A and B: Participants observed the consequences of the proposed plan.}
    \end{minipage}
    \begin{minipage}[t]{0.48\linewidth}
        \centering
        \includegraphics[width=\linewidth]{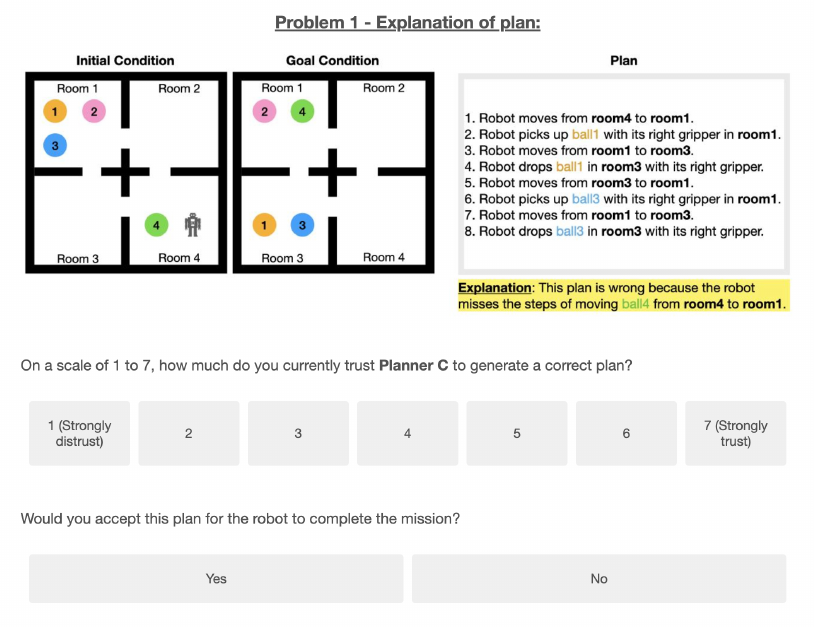}
        \caption{Planner C: Participants received an explanation of the plan.}
        \label{fig:during_intervention_c}
    \end{minipage}
\end{figure}

\begin{figure}[ht]
    \centering
    \begin{minipage}[t]{0.48\linewidth}
        \centering
        \includegraphics[width=\linewidth]{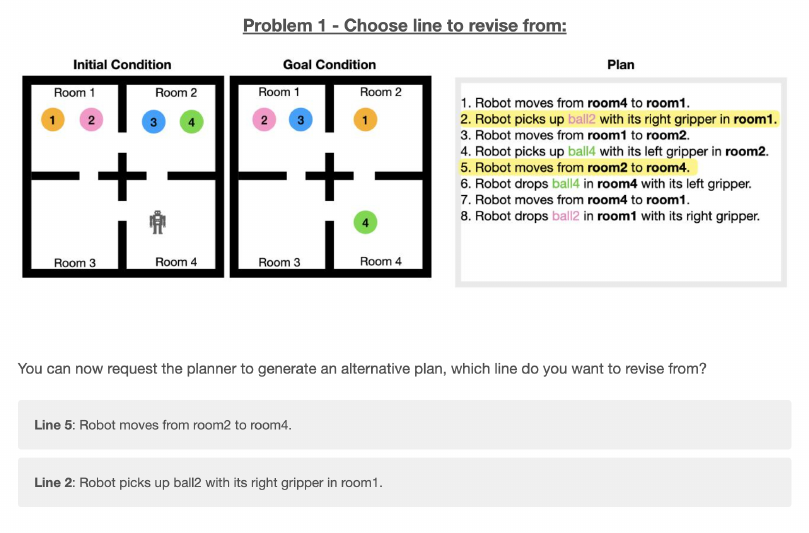}
        \caption{Participants reviewed the problem (initial and goal state) and plan with two lines highlighted. The text boxes from left to right present two choices of plan refinement (where the refinement starts)}
    \end{minipage}
    \begin{minipage}[t]{0.48\linewidth}
    \centering
    \includegraphics[width=\linewidth]{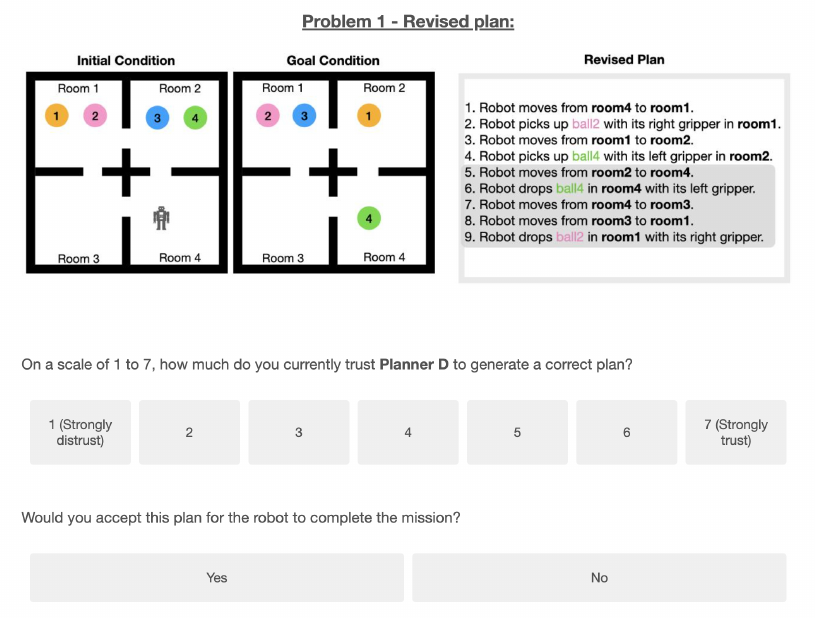}
    \caption{Participants received the refined plan.}
\end{minipage}
    \caption{Planner D.}
    \label{fig:during_intervention_d}
\end{figure}

\paragraph{During Intervention.}
Each planner involved a specific type of intervention, as detailed in \Cref{fig:during_intervention_ab,fig:during_intervention_c,fig:during_intervention_d}.

\subsection{Propensity to Trust Scale}\label{appendix:propensity_to_trust_scale}

At the end of each session, participants rated their agreement with statements about the planner they just interacted with. For example, at the end of a session with Planner A, they were asked: ``\textit{Please indicate your level of agreement with the following statements about Planner A:}".

\begin{table}[ht]
\centering
\label{tab:trust_scale}
\resizebox{\textwidth}{!}{
\begin{tabular}{lccccc}
\toprule
 & \textbf{Strongly Disagree} & \textbf{Somewhat Disagree} & \textbf{Neither Agree Nor Disagree} & \textbf{Somewhat Agree} & \textbf{Strongly Agree} \\
 \textbf{Statement} & \textbf{(1)} & \textbf{(2)} & \textbf{(3)} & \textbf{(4)} & \textbf{(5)} \\
\midrule
I usually trust Planner A until there is a reason not to. & $\circ$ & $\circ$ & $\circ$ & $\circ$ & $\circ$ \\ 
For the most part, I distrust Planner A. & $\circ$ & $\circ$ & $\circ$ & $\circ$ & $\circ$ \\ 
In general, I would rely on Planner A to assist me. & $\circ$ & $\circ$ & $\circ$ & $\circ$ & $\circ$ \\ 
My tendency to trust Planner A is high. & $\circ$ & $\circ$ & $\circ$ & $\circ$ & $\circ$ \\ 
It is easy for me to trust Planner A to do their job. & $\circ$ & $\circ$ & $\circ$ & $\circ$ & $\circ$ \\ 
I am likely to trust Planner A even when I have little knowledge about it. & $\circ$ & $\circ$ & $\circ$ & $\circ$ & $\circ$\\
\bottomrule
\end{tabular}}
\caption{Propensity to trust scale for Planner A session. The planner label is replaced with B, C, or D in other sessions.}
\end{table}

\begin{figure}
    \centering
    \includegraphics[width=\linewidth]{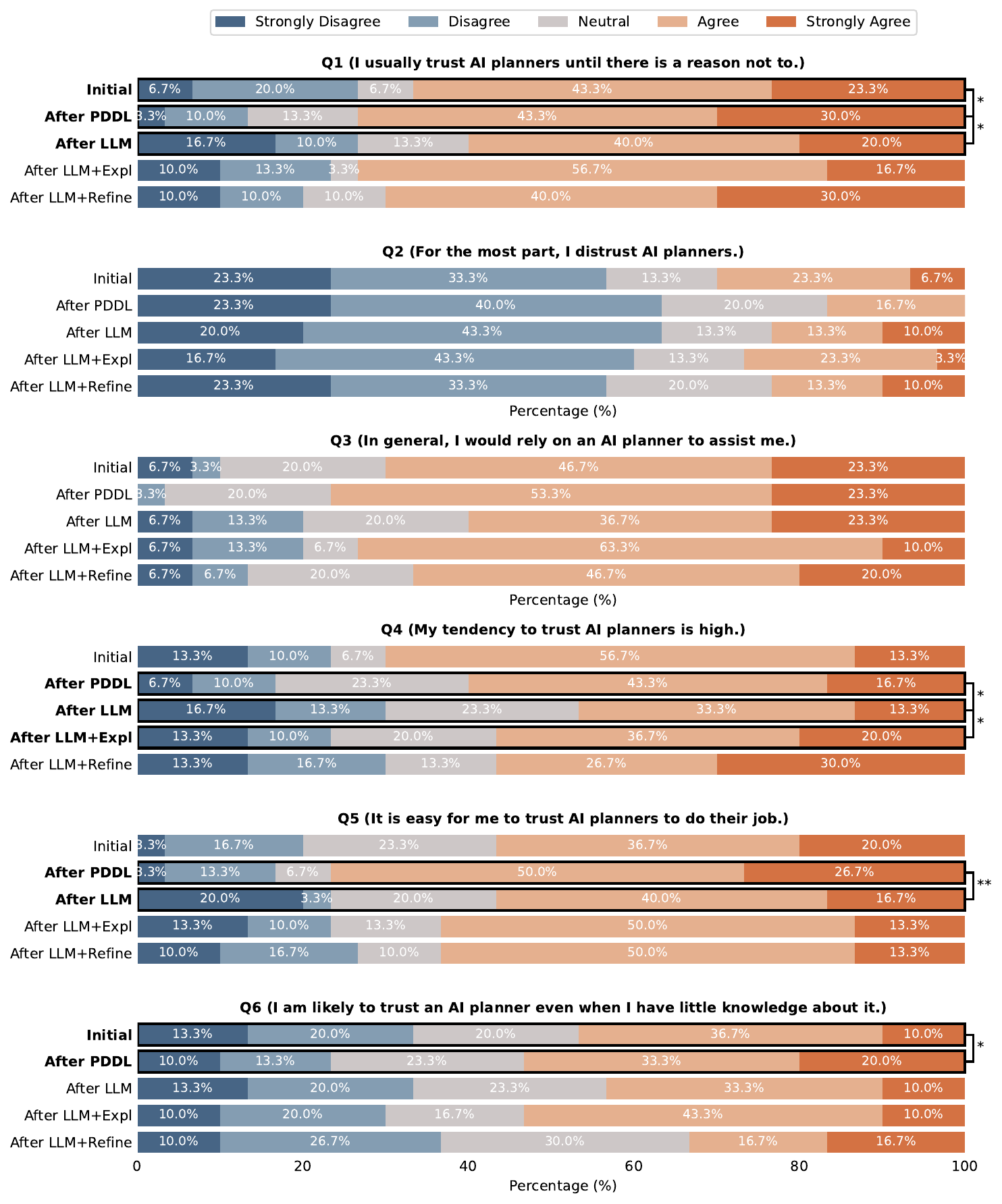}
    \caption{Complete Propensity to trust scale result}
    \label{fig:propensity_full}
\end{figure}

The results for the complete propensity to trust scale is presented in \Cref{fig:propensity_full} for reference.

\end{document}